\documentclass{article}
\usepackage{kr}

\usepackage{times}
\usepackage{soul}
\usepackage{url}
\usepackage[hidelinks]{hyperref}
\usepackage[utf8]{inputenc}
\usepackage[small]{caption}
\usepackage{graphicx}
\usepackage{amsmath}
\usepackage{amsthm}
\usepackage{booktabs}
\usepackage{algorithm}
\usepackage{algorithmic}
\usepackage{amssymb}
\usepackage{amsfonts}
\usepackage{bbm}
\usepackage{booktabs}
\usepackage{xcolor}
\usepackage{graphicx}
\usepackage{amsmath}
\usepackage{tablefootnote}
\usepackage{booktabs}
\usepackage{changepage}   
\usepackage{subcaption}
\usepackage{multirow}
\usepackage{cleveref}
\usepackage{adjustbox}
\usepackage{orcidlink}

\usepackage[dvipsnames]{xcolor}
\definecolor{Mycolor1}{HTML}{117a65}

\title{Validating Political Position Predictions of Arguments}

\author{%
Jordan Robinson\textsuperscript{1,2\orcidlink{0009-0001-6828-3524}}\and
Angus R. Williams\textsuperscript{2\orcidlink{0009-0004-3136-3768}}\and
Katie Atkinson\textsuperscript{1,2\orcidlink{0000-0002-5683-4106}}\and
Anthony G. Cohn\textsuperscript{2,3\orcidlink{0000-0002-7652-8907}} \\
\affiliations
$^1$University of Liverpool\\
$^2$The Alan Turing Institute\\
$^3$University of Leeds\\
}

\begin{document}

\maketitle

\begin{abstract}
Real-world knowledge representation often requires capturing subjective, continuous attributes -- such as political positions -- that conflict with pairwise validation, the widely accepted gold standard for human evaluation. We address this challenge through a dual-scale validation framework applied to political stance prediction in argumentative discourse, combining pointwise and pairwise human annotation. Using 22 language models, we construct a large-scale knowledge base of political position predictions for 23,228 arguments drawn from 30 debates that appeared on the UK politicial television programme \textit{Question Time}. Pointwise evaluation shows moderate human--model agreement (Krippendorff's $\alpha=0.578$), reflecting intrinsic subjectivity, while pairwise validation reveals substantially stronger alignment between human- and model-derived rankings ($\alpha=0.86$ for the best model). This work contributes: (i) a practical validation methodology for subjective continuous knowledge that balances scalability with reliability; (ii) a validated structured argumentation knowledge base enabling graph-based reasoning and retrieval-augmented generation in political domains; and (iii) evidence that ordinal structure can be extracted from pointwise language models predictions from inherently subjective real-world discourse, advancing knowledge representation capabilities for domains where traditional symbolic or categorical approaches are insufficient.

\end{abstract}

\section{Introduction}

This paper addresses the challenge of validating large-scale, pointwise large language model (LLM) predictions for subjective continuous variables, specifically political position scoring in argumentative discourse. While pointwise human annotation is scalable, humans are cognitively ill-equipped for precise, pointwise judgements~\cite{mosteller1977data,likert1932technique}. Research has long demonstrated that humans excel at comparative judgments but struggle with absolute positioning on continuous scales~\cite{thurstone_1927,Tarlow31122021}. Pairwise comparison -- where annotators judge which of two items has more of some attribute -- aligns better with human cognitive capabilities~\cite{10.1093/biomet/31.3-4.324,10.2307/2347562}, even though it is significantly more expensive: validating $n$ items requires $O(n^2)$ comparisons rather than $O(n)$ pointwise judgments, making full pairwise validation prohibitive at scale. We propose a dual-scale validation framework that combines both approaches: pointwise validation to identify political arguments and pairwise validation to assess the relative ordering of political positions. 



Grounded in a knowledge base of 23,228 argumentative discourse units (ADUs)~\cite{Peldszus_and_Stede_ADUs} extracted from 30 BBC \textit{Question Time} debates, we employ 22 LLMs to predict the political positions of arguments along the left--right wing spectrum. Human validation is conducted in two stages using over 1,500 crowdworkers. The resulting knowledge graph integrates formal argumentative relations with political position predictions, enabling fine-grained analysis of political discourse and downstream applications, such as graph-based retrieval-augmented generation (RAG). 

Argument(ation) mining and political science have largely evolved in isolation until recently, with early work on political argumentation mining published by Lippi and Torroni~ \shortcite{Lippi_Torroni_2016}. Argument mining~\cite{lawrence-reed-2019-argument,10.1145/2850417} focusses on identifying argumentative structure -- such as premises, conclusions, relations like support and attack, and argumentation schemes -- paying no attention to an argument's political sentiment. 

This separation leaves a gap: we still lack large-scale, structured resources that jointly represent \textit{what} is being argued and \textit{where} those arguments fall on the political spectrum. Existing approaches struggle to support granular analyses of political discourse, such as how ideological positions propagate through argumentative structures or how political bias manifests at the ADU-level.

Our work directly addresses this gap. We introduce the first large-scale knowledge base that unifies locutions and their corresponding ADUs with political position predictions, enabling political stance analysis at the level of the atoms of individual arguments, rather than entire texts or speakers. Taking 22 LLMs and validating their predictions through a novel dual-scale human annotation methodology, we show that scalable pointwise model predictions can be meaningfully aligned with human comparative judgements. Thus, we extend the literature on political argumentation by combining structured argumentation corpora with validated political stance predictions. The resulting resource\footnote{Code and the containerised knowledge base are available on GitHub: \url{https://github.com/anonymous-argumentation/Validating-Political-Position-Predictions-of-Arguments}.} opens new research strands in political argumentation, computational social science and graph-based RAG, supporting analyses that were previously infeasible due to the lack of structured, politically annotated argumentative data.





\section{Related Work}

\paragraph{LLMs as Evaluators of Language Outputs.}
Prior to recent advancements in LLMs, evaluation of natural language generation primarily relied on automated metrics, such as BLEU~\cite{papineni-etal-2002-bleu}, ROUGE~\cite{lin-2004-rouge}, and later embedding-based approaches including BERTScore~\cite{zhang2020bertscoreevaluatingtextgeneration}, BARTScore~\cite{yuan2021bartscoreevaluatinggeneratedtext} and GPTScore~\cite{fu2023gptscoreevaluatedesire}. While effective for supervised tasks with reference outputs, these metrics are limited in their ability to assess open-ended, normative, or context-dependent language, as they rely on surface-level similarity or static semantic representations.

The emergence of powerful LLMs capable of nuanced language understanding and reasoning in unsupervised tasks has led to their adoption as evaluators, or ``judges'', of generated text. Recent work demonstrates that LLMs can reliably assess outputs produced by both humans and other models across a range of tasks~\cite{chen2023phoenixdemocratizingchatgptlanguages,zhang2023huatuogpttaminglanguagemodel,chen2024huatuogptiionestagetrainingmedical,wang-etal-2024-cmb,chen-etal-2024-humans}, including instruction following, question answering, and complex domain-specific reasoning~\cite{huang-chang-2023-towards,zhao2025surveylargelanguagemodels}. This shift reflects a broader trend in which generative models are used not only for evaluation but also for prediction and classification tasks that require interpreting subtle semantic and pragmatic cues.

\paragraph{Political Position Prediction with LLMs.}
Within political science and computational social science, language models have been used for political stance prediction at various granularities. For example, models were tasked with the prediction of politicians' political leanings using different ideological axes, such as gun control~\cite{wu2023largelanguagemodelsused}. GPT 3 and 4 were prompted to estimate the probability of a sentence being either \textit{conservative} or \textit{liberal} across political party manifestos, taking the sentence-level average as an analogue for the mean position of each document~\cite{Ornstein_Blasingame_Truscott_2025}. Multiple models were used to predict the political stance of sentences, using a scale from 0 (\textit{extremely left}-wing) to 100 (\textit{extremely right}-wing), within sets of Tweets, manifestos, and policy speeches across ten different languages~\cite{Le_Mens_Gallego_2025}. 

Despite differences in scale and domain, this body of work predominately adopts pointwise model evaluation approaches, in which models assign absolute ideological scores or labels that are subsequently compared against aggregate human annotations. As we discuss below, such approaches may obscure systematic differences in judgement and place substantial cognitive demands on both human annotators and models. 

\paragraph{Pairwise Comparison and Preference-Based Evaluation.}
A large body of psychological and decision-theoretic research has shown that pairwise comparison is cognitively simpler and more reliable than absolute rating for subjective judgement tasks~\cite{thurstone_1927}. This insight underpins preference learning~\cite{furnkranz_preferencelearning}, which has become central to recent advances in LLM training. In particular, Reinforcement Learning from Human Feedback (RLHF) relies on pairwise human preferences to train reward models, with Bradley--Terry-style models commonly used to infer latent strength scores from comparison data~\cite{cristiano2017_deeprl,ouyang2022_humanfeedback}. 

In contrast, existing work on LLM-based political position prediction has largely relied on pointwise scores rather than direct comparative judgements~\cite{Le_Mens_Gallego_2025}. Recent studies have shown that aggregate correlation metrics can mask systematic disagreements between model outputs and human annotations, particularly when human uncertainty is high~\cite{elangovan2025_beyondcorrelation}. Accordingly, chance-corrected agreement measures, such as Krippendorff's $\alpha$, have been argued to provide a more appropriate basis for LLM--human comparison, as they explicitly account for agreement expected by chance~\cite{haldar2025_ratingroulette}. These findings motivate the use of pairwise, preference-based evaluation for political stance judgements.

\paragraph{Political Argumentation Resources.} 
Existing political argumentation and stance detection datasets predominantly provide categorical annotations. Stance detection resources typically label texts as \textit{for} or \textit{against} a target or along a discrete left--right scale~\cite{sim2013_politicalspeeches,mohammad2016_semeval}, while argument mining datasets focus on identifying argumentative components and relations without explicitly modelling ideological position~\cite{Lippi_Torroni_2016,Menini_Cabrio_Tonelli_Villata_2018,haddadan-etal-2019-yes,visser2020argumentation,mestre2021_marg,ijcai2022p575}. Continuous measures of political stance and comparative judgements between arguments remain relatively unexplored, particularly in settings that combine ideological positioning with structured argumentative content. 



\section{Knowledge Base Construction}
\label{sec:kb}

\begin{figure*}[h]
    \centering
    \includegraphics[width=1\linewidth]{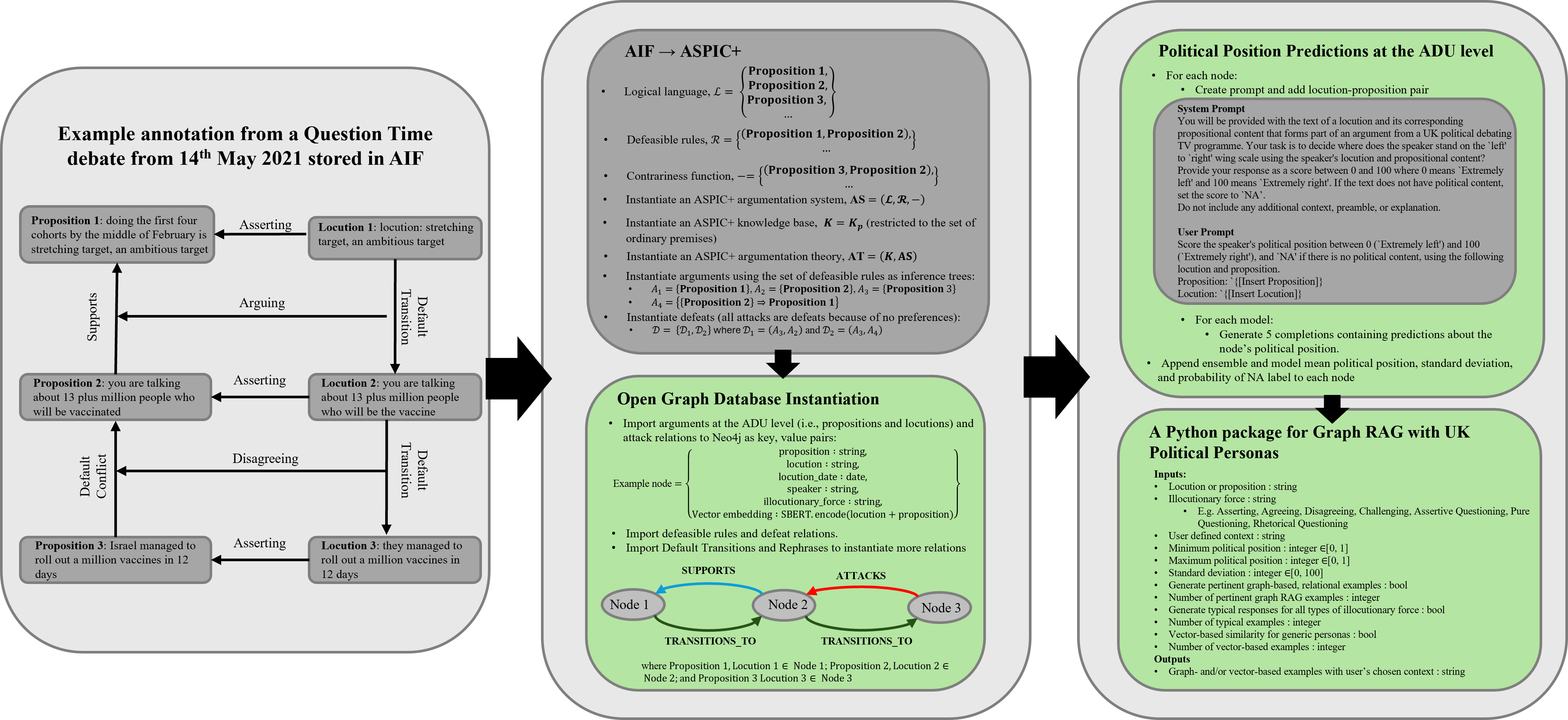}
    \caption{Overview of the methodology used to instantiate a structured argumentative knowledge base containing political positions predictions for all arguments. Green boxes indicate components developed specifically for this study.}
    \label{fig:pipeline_methodology}
\end{figure*}

Figure \ref{fig:pipeline_methodology} provides an overview of the pipeline we have developed for knowledge base instantiation, prediction of political stances, and application via graph-based RAG. We explain each part of the pipeline below.

\subsection{Data Sources and Preprocessing}
We constructed a knowledge base from 30 BBC Question Time debates\footnote{\url{https://corpora.aifdb.org/qt30}} that were previously annotated for arguments by expert annotators with an inter-annotator agreement (Combined Argument Similarity Score) of 0.56~
\cite{db732e60cc054949a08a9f37d8767690}. 
The annotated public corpora, represented in the Argument Interchange Format (AIF)~\cite{aif1,aif2}, were converted into an ASPIC$^{+}$ argumentation theory following Bex et al.~\shortcite{10.1093/logcom/exs033}, restricting the framework to ordinary premises and defeasible inference rules without preferences.

Arguments, defeat relations, and AIF dialogue transition structures were instantiated in a Neo4j graph database~\cite{noauthororeditorneo4j}. We adopt the ASPIC$^{+}$ framework because it provides a formally rigorous bridge between natural-language argument structure and Dung-style abstract semantics~\cite{aspic,aspic_tutorial}, supporting principled evaluation of argument acceptability. Granularity is important for graph-based RAG systems emulating political personas, which require ideologically aligned retrieval and coherent argumentative generation. Constructing personas with internally consistent and politically aligned argument sets is left for future work.

The resulting knowledge base contains 23,228 arguments, consisting of locution-proposition pairs, and 50,905 relations comprised of support, attack, rephrase and chronological transition. 

\subsection{LLMs as Judges of Political Position}

We used 22 LLMs to predict the political positions of arguments in the knowledge base, visible in ~\Cref{meth:tab:model_and_ensemble}. 

Building on~\cite{Le_Mens_Gallego_2025}, we extended sentence-level political stance predictions to both utterances and their corresponding ADU, interchangeably referred to as locution-proposition pairs or arguments from now on. Each model assigned a political stance score to all locution-proposition pairs on a 0-100 scale (left-right), or ``NA'' if an argument lacked political content. An example prompt is shown in Figure \ref{fig:pipeline_methodology}.

Each model scored every node in the graph five times to account for output variability even under near-deterministic settings (i.e., temperature set to 0, top$\_$p equal to 0.1, and fixed seed for models that allowed parameter configuration). Prompting was executed using Golem \cite{robert_blackwell_2024_14035711} to ensure reproducible prompts with consistent parameter configurations at scale. We derived summary statistics for political position scores and the probability of an argument being labelled apolitical (or ``NA'') across repetitions, and stored these as properties of nodes in the knowledge base.

\subsection{Ensemble Construction and Aggregation}\label{subsec:ensemble-creation}

We constructed three ensembles of model predictions, each designed to isolate different aspects of model behaviour and robustness and provide additional insights in evaluation. Summary statistics for political position score and probability of NA were calculated from the set of all individual predictions for all models in an ensemble, and stored as properties for each node in addition to individual model predictions and summary statistics.

\subsubsection{Ensemble 1 [$E_1$]: All ($n=22$).} A general-purpose baseline capturing average model behaviour across all models.

\subsubsection{Ensemble 2 [$E_2$]: Reasoning Models ($n=5$).} Models with explicit reasoning or chain-of-thought capabilities, to establish whether models designed to show their reasoning steps produce predictions more aligned with human judgements.

\subsubsection{Ensemble 3 [$E_3$]: High-Confidence Models ($n=12$).} 
Models with high-confidence judgements, defined by the number of valid political stance predictions between 0-100 exceeding the number of NA predictions. Introduced after initial results indicated that smaller models were incapable of classifying arguments as NA, even for samples which were apolitical.\footnote{E.g. some models assigned a score to nodes which contained a locution such as ``I agree'' which possessed no political sentiment.}
\begin{align}
    E_3 = \{m \in E_1 \; : \; |\hat{y}_{pol}(m)| > |\hat{y}_{apol}(m)|\}
\end{align}
\noindent where $m$ is a model, $\hat{y}_{pol}(m)$ is the number of arguments predicted as political by $m$, and $\hat{y}_{apol}(m)$ is the number arguments predicted NA by $m$.

\begin{table}[h]
\centering
\begin{adjustbox}{max width=\columnwidth}
    \begin{tabular}{llccc}
    \toprule
        \textbf{Model} & \textbf{Official Name} & \textbf{$E_1$} & \textbf{$E_2$} & \textbf{$E_3$} \\
        \midrule
        Claude 3.5 Haiku & \texttt{claude-3-5-haiku-20241022} \cite{anthropic_claude__haiku_2024}& \checkmark & & \\
        Claude 3.7 Sonnet & \texttt{claude-3-7-sonnet-20250219} \cite{anthropic_claude_sonnet_2024}&\checkmark & \checkmark & \checkmark \\
        DeepSeek-R1 & \texttt{deepseek-r1-0528} \cite{Guo_2025}&\checkmark & \checkmark & \checkmark \\
        DeepSeek-V3 &\texttt{deepseek-v3-0324} \cite{deepseekai2025deepseekv3technicalreport}&\checkmark & & \checkmark \\
        Gemini 1.5 Pro & \texttt{gemini-1.5-pro-002} \cite{geminiteam2024gemini15unlockingmultimodal}&\checkmark & & \checkmark \\
        Gemini 2.5 Flash & \texttt{gemini-2.5-flash-preview} \cite{comanici2025gemini25pushingfrontier}&\checkmark & & \checkmark \\
        GPT 3.5 Turbo &\texttt{gpt-3.5-turbo-0125} \cite{openai_gpt35_0125}&\checkmark & & \\
        GPT 4 Turbo &\texttt{gpt-4-turbo-2024-04-09} \cite{openai_gpt4_turbo_20240409}&\checkmark & & \checkmark \\
        GPT 4o &\texttt{gpt-4o-2024-08-06} \cite{openai_gpt4o_20240806}&\checkmark & & \checkmark \\
        GPT 4o Mini & \texttt{gpt-4o-mini-2024-07-18} \cite{openai_gpt4o_mini_20240718}& \checkmark & & \\
        GPT 4.5 &\texttt{gpt-4.5-preview-2025-02-27} \cite{openai_gpt45_preview_20250227}&\checkmark & & \checkmark \\
        Grok 2 & \texttt{grok-2-1212} \cite{grok-2}&\checkmark & & \checkmark \\
        Llama 3.1:8b &\texttt{llama3.1:8b} \cite{grattafiori2024llama3herdmodels}&\checkmark & & \\
        Llama 3.1:405b &\texttt{llama-3.1-405b-instruct} \cite{grattafiori2024llama3herdmodels}&\checkmark & & \\
        Llama 3.2:3b &\texttt{llama3.2:3b} \cite{llama3.2}&\checkmark & & \\
        Llama 3.3:70b &\texttt{llama-3.3-70b-instruct} \cite{llama3.3}&\checkmark & & \\
        Llama 4 Maverick &\texttt{llama-4-maverick} \cite{llama4}&\checkmark & & \checkmark \\
        Mistral:7b &\texttt{mistral:7b} \cite{jiang2023mistral7b}&\checkmark & & \\
        o3 Mini & \texttt{o3-mini-2025-01-31} \cite{o3mini}&\checkmark & \checkmark & \\
        Phi 4 & \texttt{microsoft/phi-4} \cite{abdin2024phi4technicalreport}&\checkmark & & \checkmark \\
        Qwen 3 &\texttt{qwen3-235b-a22b} \cite{yang2025qwen3technicalreport}&\checkmark & \checkmark & \checkmark \\
        Qwen QwQ &\texttt{qwq-32b} \cite{qwen-qwq}&\checkmark & \checkmark & \\
        \midrule
        &\textit{Total} & 22 & 5 & 12 \\
        \bottomrule
    \end{tabular}
    \end{adjustbox}
    \centering
    \caption{Models included in each ensemble.}
    \label{meth:tab:model_and_ensemble}
\end{table}



\section{Human Annotation and Validation Design}
\label{sec:method}

In order to validate model predictions of political positions introduced in \Cref{sec:kb}, we made use of human crowdworkers to annotate arguments from the knowledge base. Crowdworkers were recruited via Prolific.\footnote{\url{https://www.prolific.com} (Accessed on 4th December 2025)} All participants recruited resided in the UK, with English as first or primary language, and some form of higher education. 

Following the structure of prompts presented to LLMs, we divided human annotation into two sequential tasks.
 
\subsection{Pointwise Binary Classification of Political Sentiment} \label{subsec:method_presence}

This task comprised pointwise binary classification as to whether an argument is political or apolitical, validating model predictions while also providing a pool of high-confidence political arguments to validate model scores on.

\subsubsection{Annotation.} We randomly sampled 1,000 arguments from our knowledge base into three buckets, based on the mean probability of NA for Ensemble 3 ($\bar{\pi}^{(E_3)}$), in order to bias the dataset towards cases of high model confidence while still including some cases of low model confidence, as shown in Table \ref{meth:tab:sampling_buckets}.
\begin{table}[h!]
    \scriptsize
    \begin{tabular}{llcl}
    \toprule
    \textbf{Bucket} & $n$ & $\bar{\pi}^{(E_3)}$ & \textbf{Interpretation} \\
    \midrule
    \texttt{H}$_{\text{pol}}$ & 400 & $\leq 0.05$ & high confidence, political \\
    \texttt{L} & 200 & $\in [0.45, 0.55]$ & low confidence \\
    \texttt{H}$_{\text{apol}}$ & 400 & $\geq 0.95$ & high confidence, apolitical \\
    \bottomrule
    \end{tabular}
    \centering
    \caption{Sampling buckets for pointwise binary classification of the presence of political sentiment.}
    \label{meth:tab:sampling_buckets}
\end{table}

All arguments obtained a majority label from three human annotations (binary labels), using a pool of 600 crowdworkers, each annotating five arguments. Two participants exhibiting invariance and atypical completion speed were flagged and replaced. We used \textit{nominal} Krippendorff's $\alpha_n$ \cite{Hayes01042007,krippendorff2018content} to measure inter-annotator agreement, accommodating partial annotator overlap and chance agreement.



\subsubsection{Confidence-Based Dataset Partitioning.} \label{subsec:meth:point_conf}

Let $\mathcal{D}^{(\text{NA})}$ denote the full dataset comprised of arguments from the \texttt{H}$_{\text{pol}}$, \texttt{L}, and \texttt{H}$_{\text{apol}}$ confidence buckets from \Cref{meth:tab:sampling_buckets}. We partitioned $\mathcal{D}^{(\text{NA})}$ into disjoint subsets based on model prediction confidence as follows:
\begin{equation}
   \mathcal{D}^{\text{(NA)}}_{\text{conf}} = \mathcal{D}^{\text{(NA)}}_{\texttt{LOW}} \cup \mathcal{D}^{\text{(NA)}}_{\texttt{H}_{\text{apol}}}, \qquad \mathcal{D}^{\text{(NA)}}_{\text{ambig}} = \mathcal{D}^{\text{(NA)}}_{\texttt{L}}.
\end{equation}
Here, $\mathcal{D}^{\text{(NA)}}_{\text{conf}}$ contained arguments that models predicted as political or apolitical with high confidence, while $\mathcal{D}^{\text{(NA)}}_{\text{ambig}}$ consisted of arguments whose political status was characterised by model uncertainty.

By construction, these confidence-based subsets formed a strict partition of the dataset. Specifically, the high-confidence and ambiguous subsets were disjoint and their union recovers the full dataset:
\begin{align}
    \mathcal{D}^{\text{(NA)}}_{\text{conf}} \cap \mathcal{D}^{\text{(NA)}}_{\text{ambig}} = \emptyset \;\; 
\text{ and } \;\; \mathcal{D}^{\text{(NA)}}_{\text{conf}} \cup \mathcal{D}^{\text{(NA)}}_{\text{ambig}} = \mathcal{D}.
\end{align}

This guaranteed that every argument was assigned to exactly one subset, ensuring complete coverage without overlap and enabling controlled comparisons between high-confidence and ambiguous cases.

\subsubsection{Evaluation.}
We evaluated presence of political sentiment using binary human majority labels and thresholded model predictions across $\mathcal{D}^{(\text{NA})}$, $\mathcal{D}^{(\text{NA})}_{\text{conf}}$, and $\mathcal{D}^{(\text{NA})}_{\text{ambig}}$.

We calculated F1 score, precision, recall, and balanced accuracy (treating human labels as ground truth), in addition to $\alpha_n$ (framing parties as two equal-weight raters), allowing us to evaluate model performance relative to human reliability baselines.


\subsection{Pairwise Comparison of Political Position}\label{subsec:method_pair}

To validate the implicit ranking of arguments from pointwise model predictions, we presented human annotators with pairs of arguments, tasking them with annotating which one was more left- or right-leaning. Model-predicted political position scores were converted into pairwise comparisons and evaluated against human pairwise judgements via rankings derived from Bradley--Terry (BT) models \cite{bradley1952} and pairwise classification performance metrics across confidence levels.

\subsubsection{Sampling Pairs.}
We sampled $n=100$ arguments that were unanimously labelled as political by human annotators from $\mathcal{D}^{(\text{NA})}_{\texttt{H}_{\text{pol}}}$, stratified according to position scores predicted by Ensemble 3. Continuous position predictions were discretised into deciles; bins corresponding to the ranges $0-10$ and $90-100$ were empty and therefore left excluded, yielding eight non-empty bins $B$. We denote the binned position score for argument $i$, repetition $a$, under model $m$ as $s^{(m)}_{i,a} \in B = \{1, \ldots, 8\}$.


In order to do efficient comparison under resource constraints, we annotated a subset of $934$ pairs drawn from the $\binom{100}{2} = 4,950$ possible argument pairings. Pair selection was guided by the mean predicted position score $\bar{s}^{(E_3)}$ from Ensemble 3. Specifically, we sampled $44$ intra-bin pairs from each bin (except for bin $8$, where only $10$ intra-bin pairs were available) and $22$ inter-bin pairs. We denote the resulting set of annotated pairs as $\mathcal{P} = \{(i_k, j_k) | k= 1, \ldots, 934\}$ where each element $(i_k, j_k)$ corresponds to an ordered pair of arguments. 


The resulting set exceeds the $n \ln n \approx 460.5$ target for pairwise connections \cite{negahban2012_ranking}, where $n=100$ for this study. We verified full connectivity of the resulting graph, and confirmed even distributions of connections by computing node connection entropy values, defined as  the Shannon entropy \cite{shannon1948} $H(v_i)$ of an item $v_i$. This is calculated as the sum of proportions of comparisons $f$ between $v_i$ and each bin $b \in B$ ($f_{i,b}$) weighted by their logarithms, shown in \Cref{eq:entropy}. 
With an ideal upper bound $\log_2(8) \approx 3$ (uniform distribution across all eight bins), the median value of the resulting distribution was $2.5$, with more than $60\%$ of scores falling within $2.2\text{--}2.8$, indicating balanced connections (lower values relate to bins with availability constraints). 
\begin{equation}\label{eq:entropy}
H(v_i) = -\sum_{b \in B} f_{i,b} \log_2(f_{i,b}) 
\end{equation}

\subsubsection{Win Matrices.}
We represent political position judgments as a hollow comparison matrix $W \in \mathbb{R}^{100 \times 100}$, where $W_{ij}$ represents aggregated win count of item $i$ over item $j$. A win indicates that $i$ was judged as more right-wing than $j$, with draws treated as half-wins, contributing $0.5$ to both $W_{ij}$ and $W_{ji}$~\cite{davidson1970}.

\subsubsection{Model Comparisons.}

For each model $m$ (22 LLMs and 3 ensembles), we constructed a dense win matrix $W^{(m)}$ from binned political prediction scores $s^{(m)}$. Models predicted political position for all possible pairs, a superset of $\mathcal{P}$. Each cell $W^{(m)}_{ij}$, representing argument $i$ versus argument $j$, was computed across all combinations of $N$\footnote{For single LLMs, $N=5$, whereas for ensembles $N=5 \times \text{ensemble size}$.} repetitions of model predictions:
\begin{equation}\label{eq:model_win}
W^{(m)}_{ij} = \sum_{a,b=1}^{N} w\left(s^{(m)}_{i,a}, s^{(m)}_{j,b}\right)
\quad \small
w(x, y) = \begin{cases}1 & x > y \\ 0.5 & x = y \\ 0 & x < y\end{cases}
\end{equation}

\subsubsection{Human Comparisons.}

We employed 936 crowdworkers across two symmetric tasks. Participants were presented with two arguments and the question \textit{``Which statement is more left-wing?''} or \textit{``Which statement is more right-wing?''}, and were tasked with selecting argument $i$, argument $j$, or ``equal''. Each pair was annotated 3 times per task (6 annotations per pair overall). We constructed win matrices as before, with draws representing a half-win, or $+0.5$.

We transposed left-framed annotations and aggregated across tasks, resulting in 3 human win matrices: inverted left-framed $W^{(H_L)}$, right-framed $W^{(H_R)}$, and aggregate human win matrices $W^{(H)}$ = $W^{(H_L)} + W^{(H_R)}$.

\subsubsection{Pairwise Modelling.} We used BT with the Iterative Luce Spectral Ranking (I-LSR) algorithm \cite{maystre2015_ilsr} provided in Python library \texttt{choix}, to model pairwise judgements.\footnote{\url{https://choix.lum.li/} (Accessed on: 12th January 2026)} This mapped comparisons onto a latent scale $\boldsymbol{\theta}$, where the probability of an outcome was given by $p(i > j) = \frac{e^{\theta_i}}{{e^{\theta_i}} + e^{\theta_j}}$. 
This process yielded a ``smoothed'' $100\times100$ probability matrix, imputing missing comparisons, and a ranking by sorting arguments in descending order of $\theta_i$, where higher values indicate more right-wing positions. We use $\mathcal{R}^{(m)}$ to represent ranking determined by model $m$.

\subsubsection{Confidence-based Dataset Partitioning.} \label{subsec:meth:pair_conf}

We assigned pairs from $\mathcal{P}$ into subsets based on the confidence of human and model (using ensemble 3) judgements, to account for uncertainty introduced across model repetitions and through the 6 multi-class human annotations per pair.

A given pair $i,j$ in $\mathcal{P}$ was assigned two confidence values ($|W_{ij}-0.5|\ge0.25 \in \{0,1\}$), for $W^{(E_{3})}$ and $W^{({H})}$, with $1$ indicating high-confidence. This provided four subsets $\mathcal{P}_{E_{3},H} \subset \mathcal{P}$ across different combinations of model and human confidence (\Cref{tab:pair_confidence}).

\begin{table}
\scriptsize
\label{tab:pair_confidence}
\centering
\begin{tabular}{ccc}
\toprule
 & $E_3$ Confident & $E_3$ Unconfident\\
\midrule
$H$ Confident & $\mathcal{P}_{1,1}$ & $\mathcal{P}_{0,1}$ \\
$H$ Unconfident & $\mathcal{P}_{1,0}$ & $\mathcal{P}_{0,0}$ \\
\bottomrule
\end{tabular}
\caption{Subsets of $\mathcal{P}$ by model and human confidence.}
\end{table}

\subsubsection{Evaluation.} 

We quantified agreement between rankings on our $n=100$ arguments using Spearman’s Footrule Distance~\cite{spearmanfr} and Kendall’s $\tau$ Distance ~\cite{kendalltau}, min-max normalised to [0,1] and inverted to represent similarity metrics, which we represent as $d_\text{footrule}$ and $d_{\tau}$ respectively. We also calculated ordinal Krippendorff's $\alpha_o$ to represent agreement. We calculate these metrics between each model ranking $\mathcal{R}^{(m)}$ and the human ranking $\mathcal{R}^{(H)}$, and use distance/agreement between aggregate human ranking $\mathcal{R}^{(H)}$ and rankings derived from the left- and right-framed annotation tasks ($\mathcal{R}^{(H_L)}$ and $\mathcal{R}^{(H_R)}$) to provide context to model results.


We calculated pairwise performance (macro-f1) of models, using human judgements as ground-truth, for all pairs receiving human annotation ($\mathcal{P}$). Labels represent a win or loss for the argument $i$ in the pair ($i,j$) as judged by model $m$, and are calculated by thresholding $W_{ij}^{(m)}$ at $0.5$.

We present results across high- and low- human and model confidence level subsets $\mathcal{P}_{E_{3},H}$ to breakdown how model performance differs across confidence levels.

To further contextualise model performance, we established two control baselines:
\begin{enumerate}
    \item \textbf{Random Baseline:} no discriminative ability, i.e. assigns equal strength parameters (zero values) to all items, resulting in uniform probability matrices of $0.5$. 
    \item \textbf{Worst-Case Baseline:} systematically inverted predictions relative to human judgments, generated by negating the human aggregate model parameters and inverting probability matrices. 
\end{enumerate}



\section{Experimental Results and Evaluation}

\subsection{Pointwise Annotation Study}\label{sec:res:pointwise}

\paragraph{Inter-Annotator Agreement.} 
Overall, $48.5\%$ of pointwise annotations were unanimous (3/3). To assess the reliability of the crowdsourced labels, we computed nominal Krippendorff's $\alpha_n$. Across the full dataset $\mathcal{D}^{\text{(NA)}}$ (n=1,000), inter-annotator agreement was low ($\alpha_n=0.305$), indicating poor agreement amongst annotators.

When restricting the analysis to the confident subset $\mathcal{D}^{\text{(NA)}}_{\text{conf}}$ (n=800), agreement increased slightly to $\alpha_n=0.317$, whereas when only ambiguous cases were considered $\mathcal{D}^{\text{(NA)}}_{\text{ambig}}$ (n=200), agreement decreased to $\alpha_n=0.259$. We note that $\alpha_n=0.483$ for unanimously-labelled arguments (n=485) and $\alpha_n=0.436$ for majority labels (n=515).

\begin{figure}[h]
    \centering
        \includegraphics[width=1\linewidth]{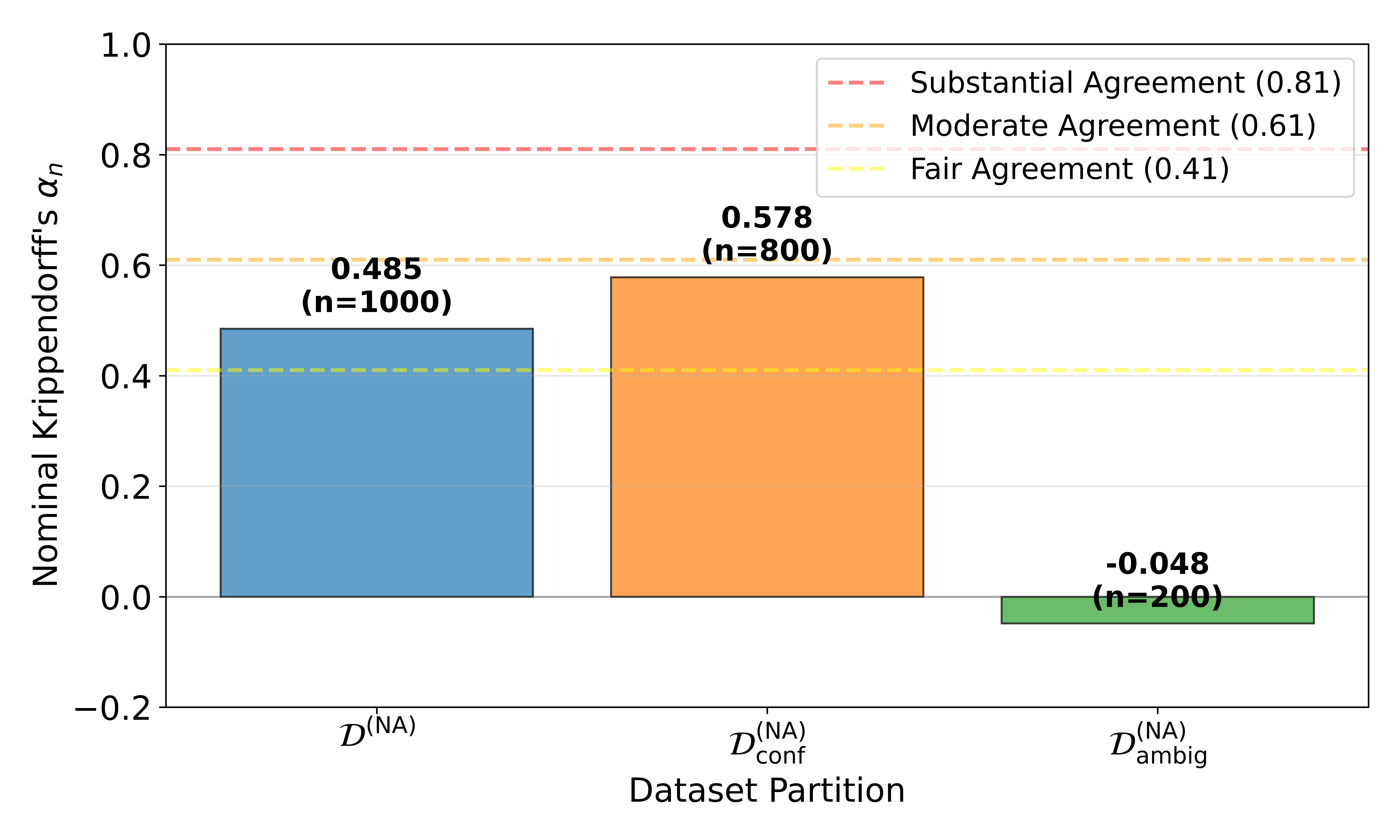}
    \caption{Nominal Krippendorff's $\alpha_n$ measuring inter-model agreement across data partitions.}
    \label{res:fig:human_and_model_agreement}
\end{figure}

\paragraph{Inter-Model Agreement.}
Figure \ref{res:fig:human_and_model_agreement} reports $\alpha_n$ measuring agreement among model predictions under different dataset partitions. When models were evaluated exclusively on the ambiguous subset $\mathcal{D}^{\text{(NA)}}_{\text{ambig}}$, agreement is at or below chance ($\alpha_n = -0.048$), indicating highly inconsistent predictions in regions of uncertainty.

When considering the full dataset $\mathcal{D}^{\text{(NA)}}$, there was a substantial increase in agreement between models, such that models exhibited fair agreement ($\alpha_n = 0.485$), with the best model agreement observed in the confident partition $\mathcal{D}^{\text{(NA)}}_{\text{conf}}$ ($\alpha_n=0.578$), approaching the moderate agreement level. 

\begin{figure}[h]
    \centering
    \includegraphics[width=1\linewidth]{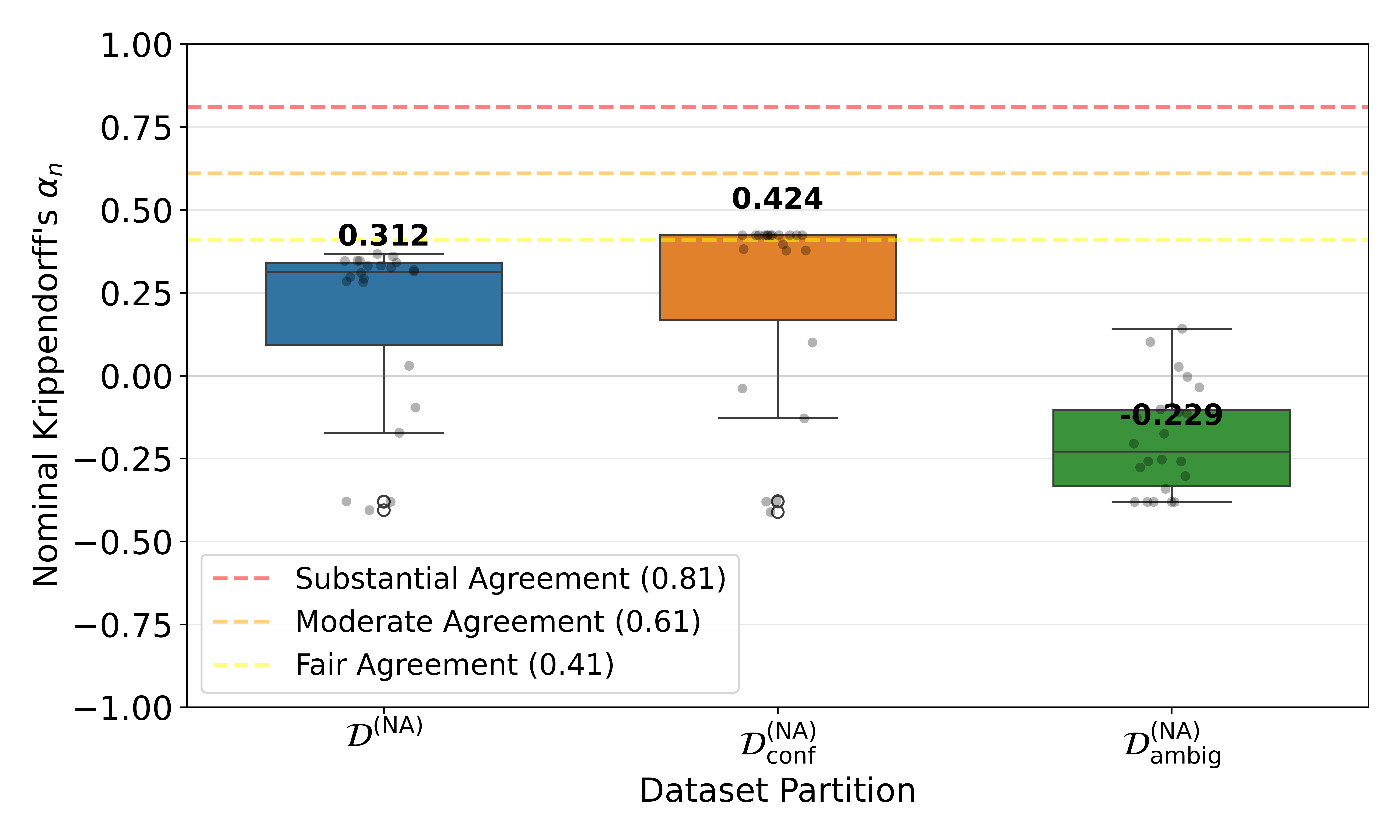}
    \caption{Distribution of human--model agreement across dataset partitions.}
    \label{res:fig:distribution-of-model-human-agreement-by-bucket}
\end{figure}

\paragraph{Human--Model Agreement.}
~\Cref{res:fig:distribution-of-model-human-agreement-by-bucket} shows the distribution of $\alpha_n$ between human annotations and model predictions across dataset partitions. Human--model agreement is highest on $\mathcal{D}^{\text{(NA)}}_{\text{conf}}$, with a median of $\alpha_n=0.424$. In contrast, agreement on $\mathcal{D}^{\text{(NA)}}_{\text{ambig}}$ is consistently negative, indicating systematic divergence between model predictions and human annotations. 

\begin{figure*}[h]
    \centering
    \includegraphics[width=1\linewidth]{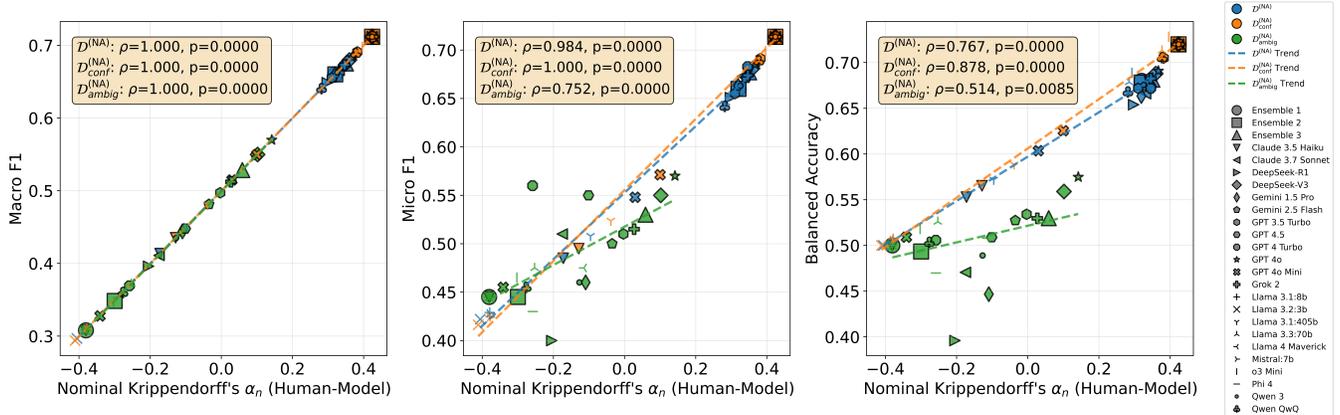}
    \caption{Relationship between human--model agreement and model performance metrics.}
    \label{res:fig:pointwise-metrics-vs-krip-alpha}
\end{figure*}

\paragraph{Model and Ensemble Performance.}
~\Cref{res:fig:pointwise-metrics-vs-krip-alpha} illustrates the relationship between human--model agreement and model performance (macro F1, micro F1 and balanced accuracy) across all models and ensembles, evaluated on $\mathcal{D}^{\text{(NA)}}$, $\mathcal{D}^{\text{(NA)}}_{\text{conf}}$, and $\mathcal{D}^{\text{(NA)}}_{\text{ambig}}$. Across all metrics, higher agreement with human judgements strongly predicts better performance. Macro F1 exhibits near-perfect correlations with agreement in both partitions and the full dataset, with similarly strong correlations observed for micro F1 and balanced accuracy. 

Performance is consistently higher on $\mathcal{D}^{\text{(NA)}}_{\text{conf}}$, demonstrating that models are most reliable on unambiguous instances. Moreover, $\mathcal{D}^{\text{(NA)}}_{\text{ambig}}$ exhibits substantial class imbalance as demonstrated by comparing macro and micro F1, likely contributing to degraded performance across all metrics and weaker agreement in this partition. These results establish human--model agreement as a robust proxy for model quality in pointwise political sentiment classification.

Beyond agreement-based analysis, we explicitly evaluated predictive performance for the three ensembles introduced in \Cref{subsec:ensemble-creation}, alongside the best performing models (\Cref{tab:ensemble_top5_all_partitions}). All three ensembles demonstrate remarkably consistent performance, with near-identical metrics across most conditions. Again, dataset partitioning had a substantial impact on performance across all ensembles and models, with results from $\mathcal{D}^{\text{(NA)}}_{\text{conf}}$ consistently outperforming both ambiguous cases and the full dataset. 
\begin{table}[h]
\centering
\begin{adjustbox}{max width=\columnwidth}
\begin{tabular}{lccccccccc}
\toprule
 & \multicolumn{3}{c}{Macro F1} & \multicolumn{3}{c}{Micro F1} & \multicolumn{3}{c}{Bal. Acc.} \\
\cmidrule(lr){2-4} \cmidrule(lr){5-7} \cmidrule(lr){8-10}
Model & $\mathcal{D}^{(\text{NA})}$ & $\mathcal{D}^{\text{(NA)}}_{\text{conf}}$ & $\mathcal{D}^{\text{(NA)}}_{\text{ambig}}$ & $\mathcal{D}^{(\text{NA})}$ & $\mathcal{D}^{\text{(NA)}}_{\text{conf}}$ & $\mathcal{D}^{\text{(NA)}}_{\text{ambig}}$ & $\mathcal{D}^{(\text{NA})}$ & $\mathcal{D}^{\text{(NA)}}_{\text{conf}}$ & $\mathcal{D}^{\text{(NA)}}_{\text{ambig}}$ \\
\midrule
Ensemble 1 & 0.660 & 0.712 & 0.308 & 0.660 & 0.714 & 0.445 & 0.680 & 0.720 & 0.500 \\
Ensemble 2 & 0.660 & 0.712 & 0.348 & 0.660 & 0.714 & 0.445 & 0.678 & 0.720 & 0.493 \\
Ensemble 3 & 0.675 & 0.712 & 0.528 & 0.677 & 0.714 & 0.530 & 0.681 & 0.720 & 0.530 \\
\midrule
GPT 4o & 0.683 & 0.712 & 0.570 & 0.685 & 0.714 & 0.570 & 0.691 & 0.720 & 0.575 \\
DeepSeek-V3 & 0.680 & 0.712 & 0.550 & 0.681 & 0.714 & 0.550 & 0.688 & 0.720 & 0.559 \\
Grok 2 & 0.673 & 0.712 & 0.512 & 0.674 & 0.714 & 0.515 & 0.682 & 0.720 & 0.530 \\
Gemini 2.5 Flash & 0.671 & 0.712 & 0.481 & 0.671 & 0.714 & 0.500 & 0.683 & 0.720 & 0.527 \\
GPT 3.5 Turbo & 0.655 & 0.691 & 0.497 & 0.655 & 0.691 & 0.510 & 0.672 & 0.705 & 0.534 \\
GPT 4.5 & 0.673 & 0.712 & 0.448 & 0.681 & 0.714 & 0.550 & 0.674 & 0.720 & 0.509 \\
Llama 4 Maverick & 0.666 & 0.712 & 0.441 & 0.666 & 0.714 & 0.475 & 0.680 & 0.720 & 0.508 \\
Gemini 1.5 Pro & 0.660 & 0.712 & 0.444 & 0.663 & 0.714 & 0.460 & 0.663 & 0.720 & 0.447 \\
Qwen 3 & 0.663 & 0.712 & 0.435 & 0.663 & 0.714 & 0.460 & 0.676 & 0.720 & 0.489 \\
Claude 3.7 Sonnet & 0.666 & 0.712 & 0.411 & 0.673 & 0.714 & 0.510 & 0.666 & 0.720 & 0.471 \\
\bottomrule
\end{tabular}
\end{adjustbox}
\caption{Ensemble and top ten performing models across $\mathcal{D}^{(\text{NA})}$, as well as $\mathcal{D}^{(\text{NA})}_{\text{conf}}$ and $\mathcal{D}^{(\text{NA})}_{\text{ambig}}$ partitions.}
\label{tab:ensemble_top5_all_partitions}
\end{table}

\paragraph{Discussion.} 
Across inter-annotator, inter-model, and human--model analyses, a consistent pattern emerges: agreement is systematically higher on $\mathcal{D}^{\text{(NA)}}_{\text{conf}}$ and degrades sharply when ambiguous instances are included. Human annotators exhibit low overall agreement, reflecting the inherent subjectivity of pointwise political sentiment annotation, while models achieve substantially higher consistency on the same confident subset. 

Both humans and models struggle on $\mathcal{D}^{\text{(NA)}}_{\text{ambig}}$, indicating a region of genuine semantic uncertainty rather than stochastic prediction noise. The convergence between human disagreement and model uncertainty provides empirical evidence that human--model agreement is a reliable indicator of model prediction quality. Excluding ambiguous instances yields marked improvements in both agreement and downstream performance, reinforcing the conclusion that pointwise model predictions can approximate human judgements of pointwise binary classification of political sentiment.

\subsection{Validating Model Predictions using Pairwise Human Annotations}

\paragraph{Inter-Annotator Agreement.} 
We measured ordinal Krippendorff's $\alpha_o$ between the inverted left-framed and right-framed annotation tasks to assess the reliability of the crowdsourced pairwise comparisons. From a carefully selected subset of 100 political arguments, we constructed 934 samples and obtained $\alpha_o=0.889$, indicating substantial agreement between framing and supporting the robustness of our pairwise annotation protocol. 


\Cref{tab:unique_labels_distribution} reports the distribution of the number of unique labels assigned per pair across annotators for the left-framed and right-framed pairwise annotations tasks. In both conditions, the majority of pairs received two distinct labels (57.3\% left-framed; 62.8\% right-framed), indicating broader annotator consensus with limited disagreement. Pairs exhibiting complete agreement -- characterised by a single unique label -- account for a substantial minority of comparisons (27.9\% and 22.5\%, respectively), while complete disagreement (three unique labels) is rare and occurred at comparable rates across framings (14.8\% left-framed; 14.7\% right-framed). The close correspondence between distributions further confirms that inter-rater agreement patterns were stable across annotation framings.
\begin{table}[h]
\scriptsize
    \centering
    \begin{tabular}{ccccc}
    \toprule
         & \multicolumn{2}{c}{Left-framed} & \multicolumn{2}{c}{Right-framed}  \\
       \cmidrule(lr){2-3} \cmidrule(lr){4-5}
      Number of unique labels & \% & n  & \% & n\\
       \midrule
       1 & 27.9 & 261 & 22.5 & 210 \\
       2 & 57.3 & 535 & 62.8 & 587 \\
       3 & 14.8 & 138 & 14.7 & 127 \\
       \bottomrule
    \end{tabular}
    \caption{Unique labels per pair across annotation tasks.}
    \label{tab:unique_labels_distribution}
\end{table}

\subsubsection{Model and Ensemble Performance, Agreement and Distance Metrics.}
\begin{figure}[h]
    \centering
    \includegraphics[width=1\linewidth]{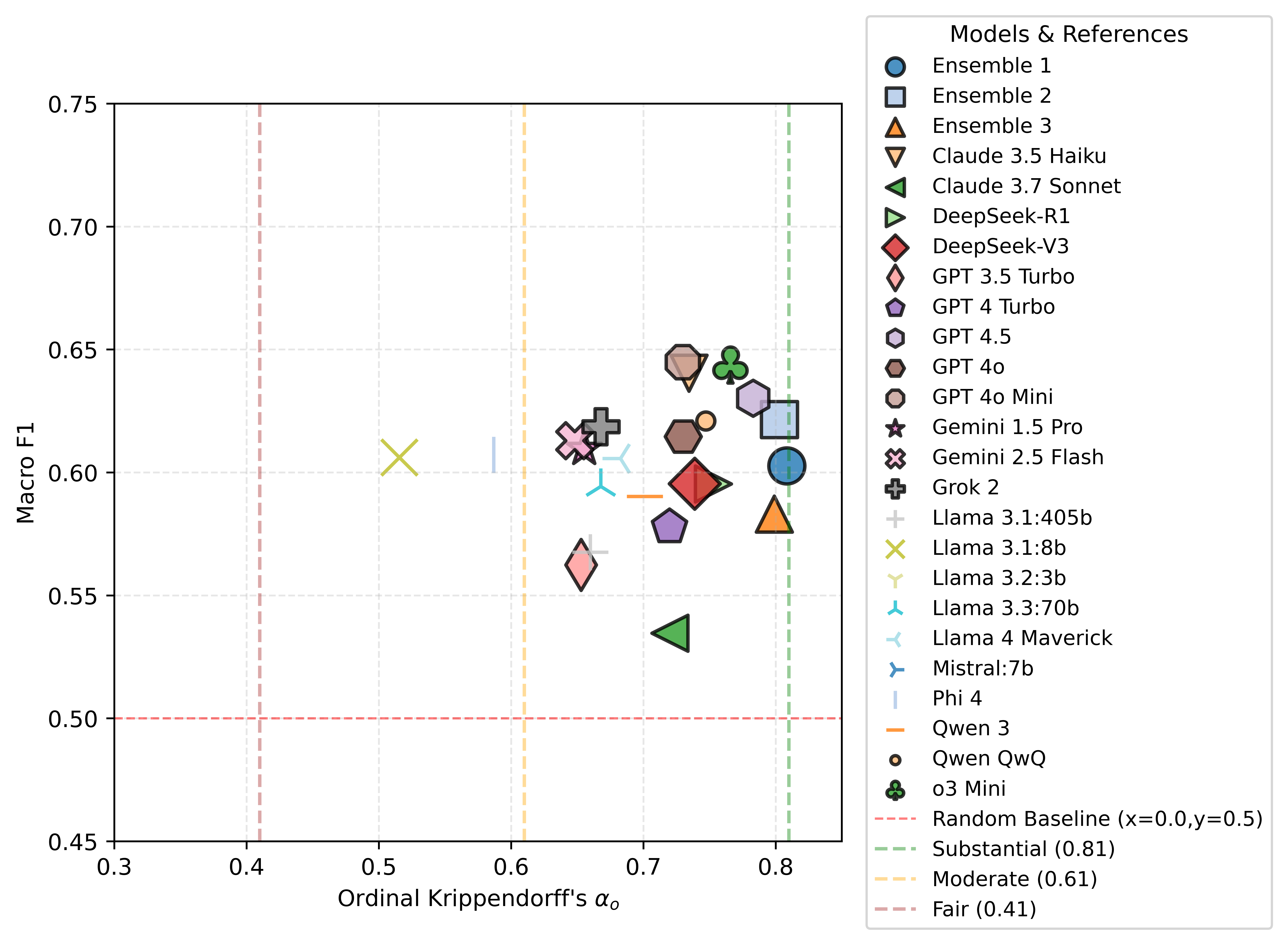}
    \caption{Macro F1 plotted against ordinal human--model agreement for all models and ensembles under $\mathcal{P}$.}
    \label{res:pairwise-full-distrib-mf1-vs-ord-kripp}
\end{figure}

\begin{figure*}[h]
    \centering
    \includegraphics[width=0.75\linewidth]{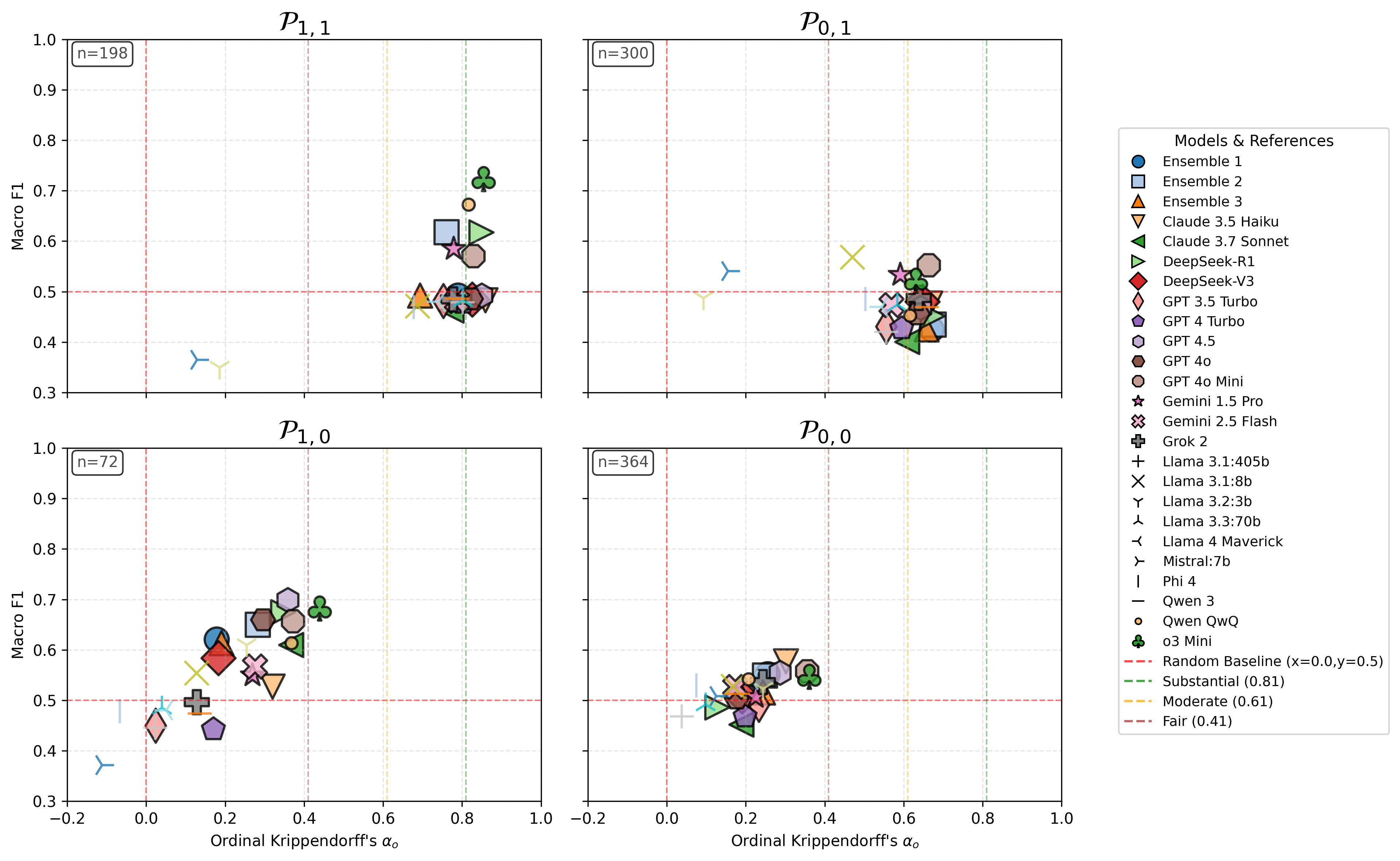}
    \caption{Macro F1 plotted against ordinal human--model agreement for all models and ensembles under four distributions: $\mathcal{P}_{1,1}$, which conditioned on high-confidence model predictions and human annotations; $\mathcal{P}_{1,0}$, which conditioned on high-confidence model predictions and low-confidence human judgements; $\mathcal{P}_{0,1}$, which conditioned on low-confidence model predictions and high-confidence human annotations; and $\mathcal{P}_{0,0}$, which conditioned on low-confidence model predictions and human annotations.}
    \label{res:fig:pairwise-sep-distrib-mf1-vs-ord-kripp}
\end{figure*}

To evaluate alignment between model-inferred political rankings and human judgements under varying levels of uncertainty, we report agreement and performance metrics over the full dataset $\mathcal{P}$ (\Cref{res:pairwise-full-distrib-mf1-vs-ord-kripp}) and over four conditional subsets (\Cref{res:fig:pairwise-sep-distrib-mf1-vs-ord-kripp}): $\mathcal{P}_{1,1}$ (high-confidence model predictions and human annotations), $\mathcal{P}_{1,0}$ (high-confidence model predictions, low-confidence human annotations), $\mathcal{P}_{0,1}$ (low-confidence model predictions, high-confidence human annotations), and $\mathcal{P}_{0,0}$ (low-confidence predictions on both sides). We report Spearman's $d_{\text{footrule}}$, Kendall's $\tau$, ordinal $\alpha_o$, and macro F1 for all ensembles and the top-performing individual models (\Cref{res:tab:pairwise-all_data_avg_score}).


Across all models, ranking agreement with humans under $\mathcal{P}$ remains substantially below human inter-annotator agreement. Individual models achieve $\alpha_o$ values which ranged from $0.52$ to $0.80$, corresponding to moderate agreement but falling short of the substantial agreement threshold ($\alpha_o=0.81$;~\Cref{res:pairwise-full-distrib-mf1-vs-ord-kripp}). This gap persists even among the strongest propriety models, indicating that inferring fine-grained political orderings from pointwise judgements remains a challenging task. 

Conditioning on high-confidence predictions markedly improves performance. Under $\mathcal{P}_{1,1}$, several models -- including GPT 4.5, o3 Mini, DeepSeek-R1 and Claude 3.5 Haiku -- approached or exceeded $\alpha_o \approx 0.85$, placing them close to the boundary of agreement between humans across the left- and right-framed tasks (\Cref{res:tab:pairwise-all_data_avg_score}). As shown in \Cref{res:fig:pairwise-sep-distrib-mf1-vs-ord-kripp}, agreement increased and variance decreased in subsets where human confidence is high, whereas low-confidence subsets exhibited substantially weaker and more variable alignment.

\begin{table}[h]
\scriptsize
\centering

\begin{adjustbox}{max width=\columnwidth}
\begin{tabular}{lcccccccc}
\toprule
 & \multicolumn{2}{c}{$d_\text{footrule}$} & \multicolumn{2}{c}{$d_\tau$} & \multicolumn{2}{c}{$\alpha_o$} & \multicolumn{2}{c}{Macro F1} \\
\cmidrule(lr){2-3} \cmidrule(lr){4-5} \cmidrule(lr){6-7} \cmidrule(lr){8-9}
Model & $\mathcal{P}$ & $\mathcal{P}_{1,1}$ & $\mathcal{P}$ & $\mathcal{P}_{1,1}$ & $\mathcal{P}$ & $\mathcal{P}_{1,1}$ & $\mathcal{P}$ & $\mathcal{P}_{1,1}$ \\
\midrule
Human (agg) & 1.000 & 1.000 & 1.000 & 1.000 & 1.000 & 1.000 & 1.000 & 1.000 \\
Human (left) & 0.904 & 0.867 & 0.934 & 0.907 & 0.974 & 0.956 & 0.827 & 1.000 \\
Human (right) & 0.886 & 0.873 & 0.922 & 0.914 & 0.966 & 0.960 & 0.790 & 1.000 \\
\midrule
Ensemble 1 & 0.726 & 0.721 & 0.810 & 0.800 & 0.808 & 0.789 & 0.603 & 0.492 \\
Ensemble 2 & 0.721 & 0.682 & 0.804 & 0.782 & 0.803 & 0.760 & 0.622 & 0.617 \\
Ensemble 3 & 0.728 & 0.657 & 0.806 & 0.756 & 0.799 & 0.692 & 0.583 & 0.492 \\
\midrule
o3 Mini & 0.684 & 0.779 & 0.779 & 0.837 & 0.766 & 0.852 & 0.644 & \textbf{0.723} \\
GPT 4.5 & 0.696 & \textbf{0.806} & \textbf{0.793} & \textbf{0.852} & \textbf{0.783} & 0.849 & 0.630 & 0.492 \\
DeepSeek-R1 & 0.691 & 0.774 & 0.781 & 0.840 & 0.753 & 0.849 & 0.595 & 0.617 \\
Qwen QwQ & 0.661 & 0.760 & 0.770 & 0.819 & 0.747 & 0.813 & 0.621 & 0.673 \\
GPT 4o Mini & 0.661 & 0.780 & 0.767 & 0.829 & 0.730 & 0.827 & \textbf{0.645} & 0.570 \\
Claude 3.5 Haiku & 0.679 & 0.785 & 0.772 & 0.842 & 0.734 & \textbf{0.858} & 0.640 & 0.483 \\
DeepSeek-V3 & \textbf{0.698} & 0.753 & 0.777 & 0.819 & 0.739 & 0.827 & 0.595 & 0.484 \\
Gemini 1.5 Pro & 0.612 & 0.743 & 0.734 & 0.806 & 0.655 & 0.778 & 0.610 & 0.585 \\
\midrule
Random Baseline & 0.333 & 0.333 & 0.500 & 0.500 & 0.000 & 0.000 & 0.500 & 0.500 \\
Worst-case Baseline & 0.000 & 0.000 & 0.000 & 0.000 & -1.000 & -1.000 & 0.000 & 0.000 \\
\bottomrule
\end{tabular}
\end{adjustbox}
\caption{Model performance metrics, by metric and dataset. Features all models that acheived top-5 performance for each metric. Bold cells indicate highest performing LLM for given column.}
\label{res:tab:pairwise-all_data_avg_score}
\end{table}

\subsubsection{Human Upper Bound.}
Human annotations established a clear upper bound on achievable performance. Aggregate human inter-annotator agreement reached $\alpha_o=1.0$ by construction, while individual left- and right-framed annotations retained extremely high agreement with the aggregate ranking ($\alpha_o\ge0.966$ for $\mathcal{P}$;~\Cref{res:tab:pairwise-all_data_avg_score}). Notably, conditioning on high-confidence judgements ($\mathcal{P}_{1,1}$) decreased the agreement slightly but increased annotation performance, yielding perfect macro F1 for both annotation conditions.

\subsubsection{Ensembles versus Individual Models.}
Under the full distribution $\mathcal{P}$, ensemble methods consistently outperformed individual models. All three ensembles achieved $\alpha_o\approx0.80$, outperforming most individual models and narrowing (but not closing) the human--model gap (\Cref{res:pairwise-full-distrib-mf1-vs-ord-kripp} and \Cref{res:tab:pairwise-all_data_avg_score}). Interestingly, this advantage diminishes under $\mathcal{P}_{1,1}$. When attention is restricted to high-confidence predictions, several individual models match or exceed ensemble performance on ranking agreement and macro F1.

\subsubsection{Ranking Agreement v. Pointwise Model Accuracy.}
A key insight from from~\Cref{res:pairwise-full-distrib-mf1-vs-ord-kripp} and~\Cref{res:tab:pairwise-all_data_avg_score} is the partial decoupling of ranking agreement and macro F1. Some models (e.g., GPT 4o Mini) achieved a strong macro F1 under $\mathcal{P}$ despite lower ranking agreement, while others prioritise ordinal consistency at the expense of pointwise accuracy. Conditioning on $\mathcal{P}_{1,1}$ sharpened this trade-off: macro F1 generally increased for confident subsets but gains in $\alpha_o$ were more pronounced, highlighting that ordinal ranking benefits more from uncertainty filtering than categorical accuracy does. 

\subsubsection{Discussion.}
Our pairwise annotation study clarifies how pointwise model-generated political judgements relate to human comparative reasoning under uncertainty. Rather than framing validation solely as agreement maximisation, our results show that confidence and aggregation play a central role in shaping the recoverable ordinal structure.

The gap between human and model agreement under the full distribution is largely attributable to uncertainty. While models exhibit only moderate alignment overall, conditioning on high-confidence judgements reveals substantial agreement with the human aggregate ranking for several models. This pattern suggests that disagreement is driven less by systematic ideological bias than by structurally ambiguous regions of the ordering that also challenge human annotators. Accordingly, model outputs encode a partially correct ordinal structure, with uncertainty concentrated in weakly constrained areas of the comparison graph.

Our findings further distinguish ordinal consistency from pointwise accuracy. Ordinal agreement and macro F1 are only weakly coupled, and improvements in one do not necessarily imply gains in the other. Confidence filtering disproportionately benefits ordinal structure, reinforcing the view that global rankings and local classifications capture distinct epistemic aspects of the task.

Aggregation effects follow a similar pattern. Ensembles improve ordinal agreement under the full distribution by stabilising rankings in ambiguous regions, but offer limited advantage once uncertainty is reduced, where strong individual models perform comparably. Finally, despite being elicited without explicit comparisons, pointwise model predictions recover a substantial portion of the ordinal information expressed in human pairwise judgements, particularly in high-confidence cases. This supports their use as a scalable proxy for human comparative annotation when exhaustive pairwise judgements are impractical.


\section{Limitations}\label{sec:limitations}
We note some limitations. 1) Pointwise political position annotations is inherently difficult for humans, as seen in substantially lower human--human agreement compared to the pairwise study, reinforcing prior findings that absolute judgments on continuous ideological scales are cognitively demanding. This limits the reliability of pointwise labels as a gold standard and motivates our dual-scale validation design, but does not eliminate the underlying subjectivity of the task. 2) Ensemble construction (esp. Ensemble 3) relies on estimating the probability that an argument is apolitical, requiring multiple model invocations per sample. At runtime, it is not known whether an argument will be unambiguous, making the approach more computationally costly and restricting it in low-latency settings. Relatedly, ensembles primarily improve performance in ambiguous regions of the comparison space; confident cases are often handled comparably well by individual models, though ambiguity itself cannot be identified prior to inference. 3) Our pairwise validation relies on discretisation of continuous model outputs and on BT regularisation choices, both of which may obscure finer-grained ordinal differences. 4) Pairwise rankings are derived from a relatively small subset of arguments (100) drawn from a single political discourse domain, which may limit generalisability to other corpora or ideological axes.

\section{Conclusions and Future Work}\label{sec:conc}


Our work shows that disagreement in pointwise political position annotation reflects human difficulty with absolute judgements rather than model unreliability. In contrast, pairwise comparison yields substantially higher agreement and reveals that pointwise LLM predictions recover meaningful ordinal political structure, particularly under high-confidence conditions. The validated subsets of our data thus delineate where political positioning can be interpreted reliably and where uncertainty is intrinsic. We release a large-scale, structured argumentation knowledge base that integrates formal argumentative structure with political position predictions at the level of individual ADUs. The graph combines attack and support relations, dialogue structure, predictions from 22 LLMs, ensemble aggregates, and uncertainty estimates. A subset of the graph has been validated by human annotators, providing a high-confidence core within a larger predictive resource and enabling fine-grained analysis beyond document- or speaker-level stance. The knowledge base can support studies on how ideology interacts with argumentative structure, how political distance relates to attack and support, and where ideological ambiguity concentrates. It also has graph-based retrieval-augmented generation, enabling retrieval of ideologically aligned and structurally coherent argument sets for downstream applications. 

Future work will 
extract political positions from ADUs in new domains, e,g, international political systems that do not have a dichotomous left-right division, and generate politically aligned personas using the shared resource.

\section*{Ethical Considerations}

This study received ethical approval from the University of Liverpool (Ref: 17359), University of Leeds (Ref: MEEC 25-001) and Alan Turing Institute (Ref: TR25-17). Human crowdworkers were recruited through the online research platform Prolific. All participants provided informed consent prior to undertaking the annotation tasks, in accordance with institutional guidelines. No personally identifiable information was collected by the research team, and participant anonymisation was managed by the Prolific platform. Participants were compensated at an average rate of £9.76 per hour across both left- and right-framed annotation tasks.

\section*{Data and Code Availability}

We release the structured knowledge base derived from 30 episodes of BBC Question Time, including arguments, their associated political position scores and relations, together with the associated Prolific-based annotation dataset and anonymised demographic metadata. The resource is provided as a database dump and Dockerfile for reproducible deployment, with supporting code available on our GitHub repository: \url{https://github.com/anonymous-argumentation/Validating-Political-Position-Predictions-of-Arguments}. All data and code are released under the MIT License.

\section*{Credit Assignment}

\textbf{Conceptualisation}: J.R., K.A., A.G.C.;
\textbf{Knowledge base construction}: J.R.;
\textbf{Predicting political positions of arguments}: J.R.;
\textbf{Human annotation study design}: J.R., A.R.W.;
\textbf{Running annotation study on Prolific}: A.R.W.;
\textbf{Analysis and data science}: J.R., A.R.W.;
\textbf{Tables}: J.R., A.R.W.;
\textbf{Visualisation}: J.R.;
\textbf{Open-source software}: J.R.;
\textbf{Supervision}: K.A., A.G.C.;
\textbf{Writing -- original draft}: J.R.;
\textbf{Writing -- review and editing}: J.R., A.R.W., K.A., A.G.C..

\section*{Acknowledgements}

The work reported in this paper was supported by funding from the Alan Turing Institute. We also acknowledge support from Microsoft Research's Accelerating Foundation Models Research programme, which provided Azure resource to run a selection of the LLMs used in the experiments that are reported in this paper.

\bibliographystyle{kr}
\bibliography{refs}

\end{document}